\documentclass[letterpaper, 10pt, conference]{ieeeconf}
\IEEEoverridecommandlockouts

\usepackage[noend, boxruled, linesnumbered] {algorithm2e}
\SetAlCapSkip{1em}
\makeatletter
\renewcommand{\algorithmcfname}{Alg.}
\renewcommand{\fnum@algocf}{\AlCapSty{\AlCapFnt\algorithmcfname\nobreakspace\thealgocf}}
\makeatother

\usepackage{svg}
\usepackage{float}
\usepackage{amsmath, array}
\usepackage{bm}
\usepackage{subcaption}
\usepackage{dsfont}
\usepackage{amssymb}
\usepackage[labelfont=bf]{caption}
\usepackage[noadjust]{cite}
\usepackage{filecontents}
\usepackage{url}

\newlength{\bibitemsep}
\newlength{\bibparskip}
\let\oldthebibliography\thebibliography
\renewcommand\thebibliography[1]{%
	\oldthebibliography{#1}%
	\setlength{\parskip}{\bibitemsep}%
	\setlength{\itemsep}{\bibparskip}%
}

\newcommand{\vvR}{v_{\text{r}}(t)}

\def\sec#1{Section \ref{sec:#1}}
\def\fig#1{Figure \ref{fig:#1}}

\def\eqn#1{(\ref{eq:#1})}

\title{\LARGE \bf Penalty-based Numerical Representation of Rigid Body Interactions with Applications to Simulation of Robotic Grasping}

\author{Michael Zechmair, Yannick Morel%
	\thanks{Michael and Yannick are with the Faculty of Psychology and Neuroscience,
		Bonnefantenstraat 2, 6211 KL Maastricht, Netherlands,
		{\tt\scriptsize \{m.zechmair, y.morel\}@maastrichtuniversity.nl}.}%
	\thanks{This research has received funding from the European Union's Horizon 2020 Framework Programme for Research and Innovation under the Specific Grant Agreement No. 785907 (Human Brain Project SGA3).}%
}

\begin{document}
	\maketitle
	\thispagestyle{empty}
	\pagestyle{empty}

	\noindent\begin{abstract}
		This paper presents a novel approach to numerically describe the interactions between rigid bodies, with a special focus on robotic grasping. Some of the more common approaches used to address such issues rely on satisfaction of a set of strict constraints, descriptive of the expected physical reality of such interactions in practice. However, application of constraint-based methods in a numerical setting may lead to problematic configurations in which, for instance, volumes occupied by distinct bodies may overlap. Such situations lying beyond the range of admissible configurations for constraint-based methods, their occurrence typically results in non-meaningful simulation outcomes. We propose a method which acknowledges the possibility of such occurrences while demoting their occurrence. This is pursued through the use of a penalty-based approach, and draws on notions of mechanical impedance to infer apposite reaction forces. Results of numerical simulations illustrate efficacy of the proposed approach.
	\end{abstract}
	\section{Introduction}\label{sec:introduction}
Over the past decade, the manufacturing industry has been undergoing a radical paradigm shift (referred to as fourth industrial revolution, \cite{schuh2017}), brought about by the development of Information and Communications Technology (ICT). The developments involved have allowed production lines to become more agile, making it possible for instance to address specialized, small product batches in a cost-effective manner (ideally, at mass production cost). A key operation involved in such production lines typically is that of assembly. Its automation by robotic means implies the necessity for the involved systems to locate, grasp, and then assemble parts in a specific configuration. The grasping operation constitutes, from a dynamical modeling perspective, a non-trivial task. It involves interactions between a robotic arm, featuring a meaningful number of Degrees of Freedom (DoFs), an object being manipulated (whose physical properties of inertia might be either partially uncertain or not straightforward to concisely represent), and an assembly onto which the grasped object must be placed or affixed. The final step involves an exchange of effort between the robotic arm, the grasped object, and the target assembly. This generally involves friction forces, which are notoriously challenging to model accurately (\cite{haessig1991}). With both the arm and the assembly usually being rigidly mounted, the operation involves a closed mechanical chain, which further complicates the process.

The interplay between these different aspects is such that the reconfiguration process of a production line involving assembly tasks oftentimes constitutes a challenging, effort-intensive proposition. While it is possible to explore this design in a numerical setting, some of the dimensions involved remain challenging to faithfully and accurately model in simulation. Grasping constitutes one of the more challenging aspects of such simulations. A number of applications have been developed to address this problem, including \texttt{GraspIt!} (\cite{graspit}) and \texttt{OpenRAVE} (\cite{openrave}). They provide a number of functions, most of them accounting for static quality of considered grasps. The involved quality metrics provide a reflection of how well a given gripper, characterised by its geometry, configuration, and actuation, is able to grasp a considered object. However, such static considerations come short of addressing the inherently dynamic nature of manipulation scenarii encountered in practice, where objects are picked, moved, and placed dynamically. Assessing quality of a grasp in such a situation requires consideration of this dynamic dimension. To that end, existing grasping applications commonly rely on numerical physics engines, such as the Open Dynamics Engine (\texttt{ODE}, \cite{smith2005}) or \texttt{Bullet} (\cite{bullet_physics_library}). The \texttt{GraspIt!} toolbox relies on \texttt{Bullet}, whereas \texttt{OpenRAVE} offers an interface allowing the use of either \texttt{ODE} or \texttt{Bullet}.

To simulate dynamic rigid body interactions, these engines rely on constraint-based methods (\cite{baraff1993non}). For each discrete time-step of the simulation, a set of constraints is defined to reflect the respective situations of the considered gripper, of the object being grasped, and of the environment. Each constraint is descriptive of interactions between two given rigid bodies, allowing to compute reaction forces emerging from contacts between them (including impacts, support, or reflecting exchange of efforts in joints). From the considered constraint is inferred an appropriate effort to be applied to the involved objects. For instance, hinge joint constraints define reactive forces that ensure the two considered bodies are kept in contact at the hinge position; they are left free to rotate around the specified axis, but other relative movements are impeded by the constraint effort. Of particular relevance to grasping are \emph{contact constraints}, which prevent objects from physically overlapping. They are implemented in a manner that is similar to the aforementioned hinge constraint. More specifically, upon detection of the emergence of one such contact, a constraint which enforces non-overlap is added to the previously considered ones (see the discussion in \cite{baraff1993non}). The constraint prevents body movement in the direction of the contact point, thus preventing overlap. Different approaches may be used to infer appropriate efforts from the defined set of constraints. One of the most commonly encountered method formulates a Linear Complementary Problem (LCP, \cite{cottle2009}), which may be solved using a number of different techniques (see \cite{anitescu_time_stepping_lcp_solver,lemke_lcp_solver}). Once evaluated, constraint forces are applied to concerned rigid bodies, thereby affecting their trajectory in the desired manner (the interested reader is referred to \cite{extending_ode_robots} for additional details).

Constraint-based approaches perform appropriately for a certain range of scenarii. However, a number of specific situations may give rise to substantial issues. For instance, rapid relative movement of objects, if simulation time steps are not carefully adjusted, may result in an effective overlap between the bodies; this overlap occurring before contact is detected and appropriate constraint forces generated to prevent the situation. A straightforward mitigating measure to such issues consists in carefully monitoring for such problems, and adjusting (that is, reducing) the simulation time-step to a degree that is commensurate with the relative velocity of concerned objects. However, such methods add considerable complexity and offer no guarantee of success (\cite{engine_simulation}). Another set of situations in which such overlaps may occur (within the context of discrete-time numerical simulation) is those involving closed mechanical chains. Grasping, when a gripper applies efforts on an object from different, opposing directions, gives rise to such situations. The issue is particularly prevalent when considering force-closure grasps (\cite{nguyen1988constructing}), for which successfully enforcing non-overlap constraints requires exceedingly small simulation time steps. In the event that an overlap ends up occurring, LCP based methods typically fail (\cite{baraff_analytical}). A detailed analysis of the performance of constraint-based rigid body simulators can be found in \cite{taylor2016analysis}, which provides an overview of most common factors leading to numerical instability in modern rigid body simulation.
\\[5pt]
\indent A number of results can be found in the literature to overcome such problems. A common approach involves accounting for the possibility of overlaps (which is largely inherent to discrete-time simulation) and computing reaction forces in such a manner that, when such situations do emerge, they only occur as short transient events, rapidly replaced with a steady-state situation in which contacts faithfully reflect the expected real-world behavior. The approach in \cite{yamane2008}, for instance, defines a set of criteria to overcome inter-body penetrations. These result in the application of a repulsive force to overlapping objects, proportional to their relative velocity at the point(s) of contact. Unfortunately, the approach implies a number of additional steps to the simulation which may prove computationally intensive, to the extent that real-time computations are in practice difficult to achieve. In addition, the defined repulsive forces can in some configurations lead to the emergence of a resonance effect, destabilizing force-closed grasps. Numerical simulation of grasping using such an approach typically requires the addition of artificial, large damping efforts diffusing excess energy (see \cite{engine_simulation}). Another existing approach involves the replacement of constraints with impulses to describe interactions, as in \cite{impulse_based}. The approach is able to account for overlap of rigid bodies. However, the reliance on impulses may give rise to a number of issues (including loss of accuracy, as discussed in \cite{impulse_based}). A manner to directly address and account for overlaps consists in treating such instances not as anomalies, but rather as errors to be reduced (or regulated). This can be pursued by defining repulsion forces in light of such errors, in a manner that concerned objects are pushed apart, as described in \cite{drumwright_penalty_based}. Penalty-based methods offer a lightweight, computationally effective approach to address simulation scenarii in which objects may find themselves intersecting with each other. Their performance has been demonstrated with implementations that allow to consider thousands of objects in real-time (\cite{sagardia_penalty}). However, in most instances, the penalty parameters (which the approach relies on) must be carefully selected for specific objects and collision scenarii, implying the need for expert knowledge and a meaningful effort overhead to achieve desired results. Furthermore, penalty-based approaches have been shown to lead to undesirable oscillating behaviors in a range of situations, as discussed in \cite{mirtich1998}. In the following, we build upon such a penalty-based perspective, proposing measures to circumvent limitations encountered with existing solutions.
\\[5pt]
\indent More specifically, when accounting for contacts in-between objects, we draw from physical considerations to infer appropriate efforts. In particular, though the objects simulated (e.g. in robotic grasping scenarii) are commonly treated as perfectly rigid, this constitutes an approximation of a more complex reality. In practice, when subjected to external efforts, objects undergo a measure of deformation (\cite{stoianovici1996critical}). In instances in which such deformations remain limited, one oftentimes overlooks their presence to facilitate proceedings. In the following instead, in situations in which bodies come in contact with one-another, we treat emerging volume overlaps between objects as zones of mechanical deformation (as may be caused by contact forces). Though not intended to provide a quantitatively faithful depiction of actual deformation phenomena, the approach affords a number of key benefits. Using models descriptive of compliant contacts to inform computation of reaction efforts, the proposed approach grounds its treatment of rigid body interactions (including unnatural configurations emerging from discrete-time simulation) in physical considerations. Building upon compliant contact models (such as those in \cite{diolaiti2005}), we propose reaction effort laws that allow to describe simulation overlaps as compliant, virtual body deformations. Thus imparting contact regions with a measure of mechanical impedance, we promote the emergence of restoring forces and moments that counteract overlaps. Further, the natural inclusion of damping within such impedance addresses the emergence of undesirable oscillatory behaviors.
\\[5pt]
\indent The proposed approach is presented in \sec{approach}. Results of illustrative numerical simulations are provided in \sec{simulation}. \sec{conclusion} concludes this paper. 
	\section{Proposed Approach}\label{sec:approach}
\baselineskip12pt
We consider the physical interaction of two or more rigid bodies. Each body is defined by its inertia properties (including mass $m\in \mathds R^{+*}$, assuming for simplicity homogeneous mass density distribution, and rotational inertia $I\in \mathds R^{3\times 3}$), a mesh describing the object's geometry as a set of vertices, and relevant friction parameters (as further discussed in \sec{simulation}). Each object's equations of motion are of the following form (\cite{tong2004lectures}),
\begin{eqnarray}
m \ddot x(t) &=& f(t),  \quad t\geqslant 0, \label{eq:dyn_trans}\\
I \dot \omega(t) &=& \tau(t) - \omega(t) \times I \omega(t), \label{eq:dyn_rot}
\end{eqnarray}
\noindent where $f(t) \in \mathds{R}^3$ is the force (in N) acting on the considered object, $\tau(t) \in \mathds{R}^3$ is the torque (in Nm), $m$ is expressed in kg, $I$ in kg$\cdot\text{m}^2$, $x(t) \in \mathds{R}^3$ represents the object's position (in m), and $\omega(t) \in \mathds{R}^3$ the object's angular velocity in $\text{s}^{-1}$. In the following, we discuss the manner in which we detect the emergence of contacts (and incidentally of overlapping object configurations), then infer appropriate reaction forces such that the overlap is mitigated.

\subsection{Overview}
\begin{figure}
	\centering
	\scalebox{1}[0.9]{\includegraphics[width=1\columnwidth]{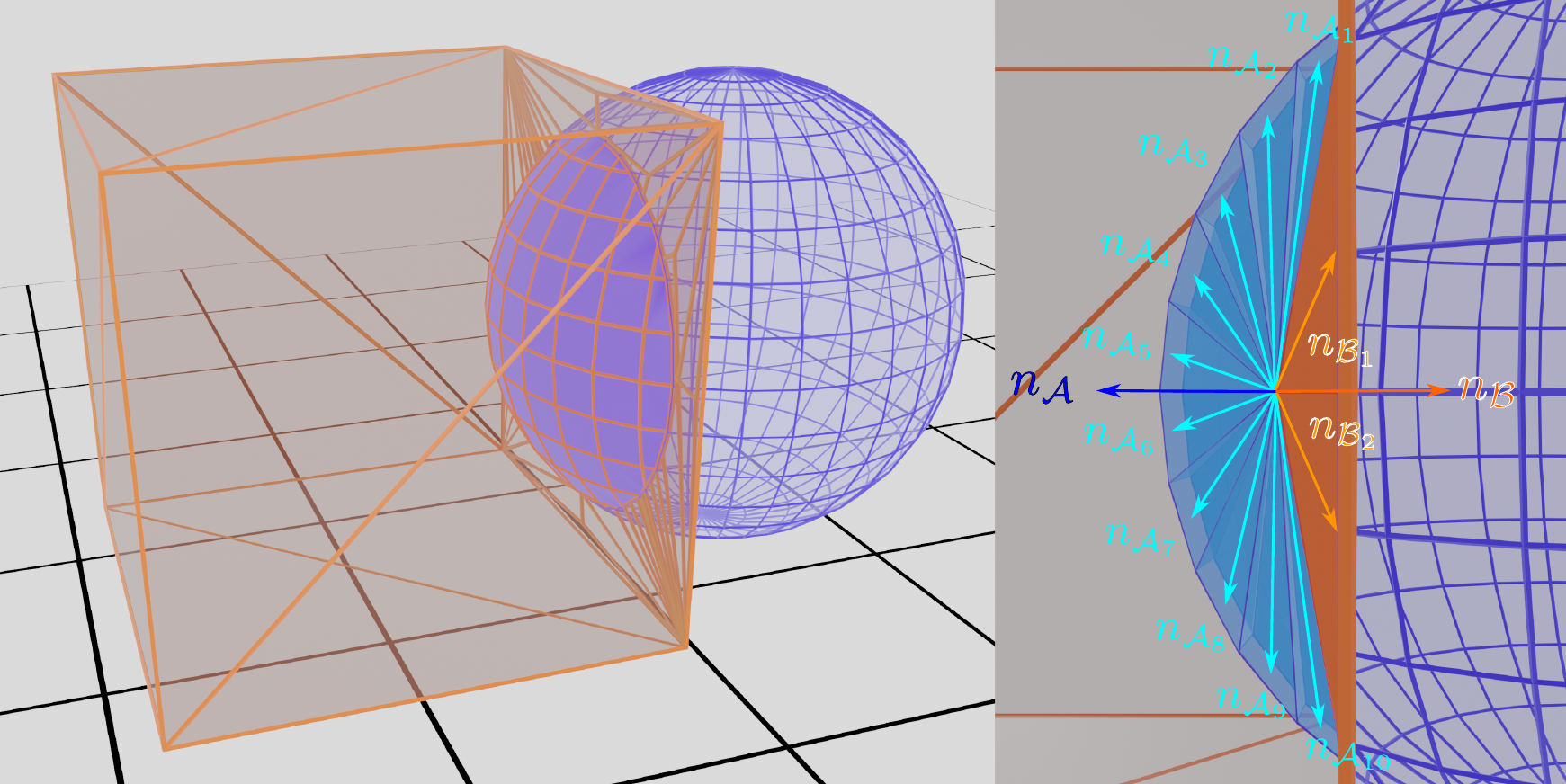}}
	\caption{Schematic representation of interaction force computation, the sphere is labeled $\mathcal{A}$, the cube $\mathcal{B}$. Upon detection of overlap, the mesh describing the overlap volume is identified (dark blue mesh, left). Evaluation of the depth of overlap is done in a discrete manner, breaking down a sample surface in a set of triangles (cyan and orange, right). The contact effort direction of forces acting from object $\mathcal{A}$ on object $\mathcal{B}$ is computed using the normals $n_\mathcal{A}$ and $n_\mathcal{B}$, $c(t)$ is the contact force application point.}
	\label{fig:overlap}
\end{figure}
Appropriate simulation of interacting rigid bodies necessarily requires that aspects related to collision detection and reaction efforts be accounted for. If relying on a penalty-based approach, such as that discussed in the following, this is typically achieved by monitoring and detecting situations in which objects overlap; that is, situations in which parts of two or more objects occupy the same space. As such situations do not arise in practice, their occurrence in simulation is undesirable, and penalty-based approaches introduce mitigating measures to overcome them. Such measures revolve around carefully designed interaction forces, developed such that overlap events are limited to transient occurrences.

The method pursued hereafter involves four successive steps. First, object collisions and volume overlaps are determined by monitoring the relative position of bodies whose dynamics are simulated. The individual meshes are intersected, and possible overlaps registered. As a direct comparison of vertices would prove exceedingly computationally expensive, this step is broken down into two stages. \texttt{\textbf 1} Object Mesh pairs in close proximity to each other are identified. \texttt{\textbf 2} Detection of a possible overlap is then performed by use of a standard Axis-Aligned Bounding Box (AABB) collision detection scheme (such as that in \cite{jimenez20013d}). \texttt{\textbf 3} Once an overlap is detected, we compute the corresponding reaction forces. Broadly speaking, the more severe the overlap, the greater the required repulsive displacement effort. Accordingly, the reaction forces we design will be made (in some measure) proportional to this volume (as is commonly the case for penalty-based techniques, see \cite{sagardia_penalty}). \texttt{\textbf 4} Direction of the resulting force vector is defined in such a manner that objects' movements decrease the overlap, and the point of application for this effort is selected. The proposed methodology is summarized in Algoritm \ref{alg:reaction_efforts}. The methodology is described in more detail hereafter.


\subsection{Characterizing the overlap}
We use the \texttt{libigl} library \cite{jacobson2016libigl} for mesh processing, which is instrumental to the proposed approach. First, we examine the meshes from selected potentially overlapping bodies and perform a standard boolean intersection operation. The result produced by \texttt{libigl} consists of a triangle mesh describing the shape of the overlap. Should the operation result in an empty set, we infer that the considered pair of bodies does not intersect and no further computations are necessary.

Should the boolean intersection provide a non-empty shape, we proceed to identifying three reaction effort parameters, the application point $c(t) \in \mathds{R}^3$, $t\geqslant 0$, the reaction effort's direction $s_{\text d}(t)\in \mathds{R}^3 $, and its magnitude $s_{\rm m}(t)\in \mathds{R}$. The point of application is assigned to the geometric center of the previously computed intersection mesh. A simple method to assign $s_{\rm m}(t)$ could involve selecting a value proportional to the scalar overlap volume measurement $v(t) \in \mathds{R}$, as is commonly done in penalty-based approaches. Such a choice leads to a proportional relationship between reaction effort and overlap volume. We propose a different approach, building upon this proportional term, in such a manner that issues commonly encountered in the use of penalty-based approaches are circumvented, as discussed in section \ref{sec:reaction_efforts}.

Assessing an appropriate $s_{\text d}(t)$, $t\geqslant 0$, requires a measure of special attention. In particular, it is necessary that collision forces be designed in such a manner that intersecting objects tend to separate from one another.
To ensure that this is the case, we approach the problem by considering the previously computed intersection mesh. More specifically, we define two sub-meshes, distinguishing, within the overlap mesh, which triangles originated from which body. For ease of exposition, we consider a case in which only two rigid bodies are overlapping, denoting them as object $\mathcal A$ and $\mathcal B$ (see \fig{overlap}). One sub-mesh is designed to include all triangles overlapping with the mesh of object $\mathcal A$, the other all triangles overlapping with the mesh of object $\mathcal B$ (as illustrated in \fig{overlap}). Then, we iterate over all triangles in subset $\mathcal A$, estimating a weight factors reflecting a triangle's contact depth and the overlap volume it represents. In greater detail, for each triangle $i$ we determine the distance vector $n_{\mathcal{A}_i}(t) \in \mathds{R}^3$ from its center point to the center of the intersection mesh $c(t)$. We then compute a weight $\omega_{\mathcal{A}_i}(t) \in \mathds{R}^+$ reflecting the scalar volume of the pyramid defined by the triangle surface as base and the point $c(t)$ as additional vertex. Once this has been performed for all triangles in the $\mathcal A$ sub-mesh, the same operation is conducted for triangles in subset $\mathcal B$. The direction of application is then computed as the following weighted sum,
%
\begin{equation}
	s_{\text d}(t) \triangleq \sum_{i=1}^{m_\mathcal{A}} \omega_{\mathcal{A}_i}(t)n_{\mathcal{A}_i}(t) - \sum_{i=1}^{m_\mathcal{B}} \omega_{\mathcal{B}_i}(t)n_{\mathcal{B}_i}(t), \quad t\geqslant0,\hspace{0.1in}
\end{equation}
\noindent where $m_\mathcal{A}$, $m_\mathcal{B} \in \mathds N$ represent the number vertices in subset $\mathcal{A}$ and $\mathcal{B}$, respectively. The value of $s_{\text d}(t)$ describes the reaction force direction vector. Efficacy of this design is illustrated in section \sec{simulation}. In the following, we discuss the design of the interaction effort magnitude, $s_{\rm m}(t)$.

%
%
%

	\subsection{Reaction Efforts} \label{sec:reaction_efforts}
%
Hereafter, we define reaction effort magnitude in such a manner that overlap in-between objects is reduced. In particular, the effort is conceived of as representative of a virtual mechanical impedance, the stiffness of which we choose proportional to the overlap volume $v(t)$, $t\geqslant0$, and add a damping term which would ideally be proportional to this volume's rate of change. However, note that there exists no direct way to compute the time derivative of $v(t)$. Using time-step differences (i.e. subtracting successive values, divided by time-step) is undesirable as it commonly proves numerically unstable (\cite{kagiwada2013numerical}). Instead, we rely on a simple second order low-pass to obtain a filtered derivative estimate. In particular, consider
\begin{eqnarray}
\ddot v_{\rm f}(t) &=& -2 \zeta \omega_{\rm n} \dot v_{\rm f}(t) + \omega_{\rm n}^2 (v(t) -v_{\rm f}(t)), \quad \dot v_{\rm f}(0)=0,\nonumber\\
                   & & \hspace{1.05in}  v_{\rm f}(0)=v(0), \quad  t\geqslant0, \label{eq:lowpass}
\end{eqnarray}
\noindent where $\zeta$, $\omega_{\rm n}\in \mathds R^{+*}$, and $\dot v_{\rm f}(t)$ provides a low-pass filtered estimate of $\dot v(t)$.

To better reflect the nonlinear (impulsive) behavior of body collision on first contact, we include a transient temporal-scale correction multiplier upon first contact. Results of numerical simulations show that use of this gain results in a substantially faster alignment of relative velocities at point of contact, and a reduced overall overlap. This gain is computed as
\begin{equation}
	\delta_{\textrm{i}} (t) \triangleq \frac{i_\textrm{v}(t)}{\sqrt{\pi}d_\textrm{v}(t)} \exp \left(-\frac{(t-t_\textrm{v}(t))^2}{d_\textrm{v}^2(t)}\right),\quad
t \geqslant 0,
\end{equation}
\noindent where
\begin{gather}
	i_\textrm{v}(t) \triangleq g_\textrm{i} (m_\textrm{A} + m_\textrm{B}) \Delta v_\textrm{c}(t), \nonumber\\
	d_\textrm{v}(t) \triangleq \frac{\textrm{d}_\textrm{t}}{\Delta v_\textrm{c}(t)}, \qquad t_\textrm{v}(t) \triangleq g_\textrm{t} d_\textrm{v}(t) - t_\textrm{c}, \nonumber
\end{gather}
\noindent $\Delta v_\textrm{c}(t) \in \mathds{R}$ is the relative speed (in m/s) of object $\mathcal{B}$ with respect to object $\mathcal{A}$ at $c(t)$ along the vector $s_{\rm d}(t)$, $\textrm{d}_\textrm{t} \in \mathds{R}^+$ is a target depth (in meters), $t_\textrm{c} \in \mathds{R}^+$ is the time instant of first contact (in $\textrm{s}$), $g_\textrm{i},$ $g_\textrm{t} \in \mathds{R}^+$ are parameters defining the magnitude and duration of applicability of gain $\delta_{\textrm{i}}(t)$, respectively. We then define the reaction efforts as follows,
\begin{align}
f_{\rm r}(t) &\triangleq \delta_{\textrm{i}} (t) \left( g_1 v(t) + g_2 \dot v_{\rm f}(t) \right) s_{\rm n}(t),\quad t \geqslant 0,\label{eq:fr}\\
 \tau_{\rm r}(t) &\triangleq (c(t)-p_{\rm o}) \times f_{\rm r}(t), \label{eq:tr}
\end{align}
where $g_1$, $g_2 \in \mathds R^{+*}$ are stiffness and damping gains (in kg/s and kg/s$^2$, respectively), $p_{\rm o}\in \mathds{R}^3$ represents the position of the considered object's center of mass, $s_{\rm n}(t)\triangleq s_{\rm d}(t)/\|s_{\rm d}(t)\|$, and $\|\cdot \|$ denotes the Euclidian norm. Note that the efforts provided by \eqn{fr}--\eqn{tr} is comparable to a linear Kelvin-Voigt contact model, descriptive of a compliant contact reaction between objects $\mathcal A$ and $\mathcal B$ (see \cite{diolaiti2005}, as well as the discussion in \cite{jain2011controlling} on modeling contact impedance). As previously alluded to, by allowing object meshes to intersect with one another, we mirror the deformation behavior expected of real-life objects at points of contact. For the purpose of numerical simulations, we built upon the models defined in \cite{diolaiti2005} and \cite{erickson2003contact} to define appropriate parameter values.
Algorithm \ref{alg:reaction_efforts} describes the manner in which we perform the computation of \eqn{fr}--\eqn{tr} when an overlap is detected.
%
%
\RestyleAlgo{boxed}
\LinesNumbered
\SetAlFnt{\footnotesize}
\IncMargin{0.4em}
\begin{algorithm}[t]
	\DontPrintSemicolon	
	\SetKwProg{Fn}{Function}{}{}
	\Fn{ComputeOverlap($\texttt{shape}_{\texttt{A}}$, $\texttt{shape}_{\texttt{B}}$)}
	{
		$\texttt{mesh}_\texttt{ov} = \texttt{bool\_intersect} \left( \texttt{mesh}_{\texttt{A}}, \texttt{mesh}_{\texttt{B}} \right)$\\
		\BlankLine\BlankLine
		\tcc{Sort triangles according to corresponding object}
		$\texttt{tri}_\texttt{A} = \texttt{extract} \left( \texttt{mesh}_{\texttt{ov}}, \texttt{mesh}_{\texttt{A}} \right)$\\
		$\texttt{tri}_\texttt{B} = \texttt{extract} \left( \texttt{mesh}_{\texttt{ov}}, \texttt{mesh}_{\texttt{B}} \right)$\\
		\tcc{Determine depth weight of individual triangles}
		$s_{\text d}(t) \leftarrow 0$\\
		\ForEach{$\texttt{v}_\texttt{A} \in \texttt{tri}_\texttt{A}$}
		{
			$\texttt{depth}_{\texttt{v}_\texttt{A}} = \| \texttt{center}\left( \texttt{tri}_\texttt{A} \right) - c(t) \|$\\
			$\omega_{\texttt{v}_\texttt{A}} \leftarrow \frac{1}{3} \texttt{surf} \left( \texttt{v}_\texttt{A} \right) \times \texttt{depth}_{\texttt{v}_\texttt{A}}$\\
			$\omega_\texttt{A} \leftarrow \left\{\omega_\texttt{A}; \omega_{\texttt{v}_\texttt{A}} \right\}$\\
			$s_{\text d} \leftarrow s_{\text d} + \omega_{\texttt{v}_\texttt{A}} n_{\texttt{v}_\texttt{A}}$\\
		}
	
		\ForEach{$\texttt{v}_\texttt{B} \in \texttt{tri}_\texttt{B}$}
		{
			$\texttt{depth}_{\texttt{v}_\texttt{B}} = \| \texttt{center}\left( \texttt{tri}_\texttt{B} \right) - c(t) \|$\\
			$\omega_{\texttt{v}_\texttt{B}} \leftarrow \frac{1}{3} \texttt{surf} \left( \texttt{v}_\texttt{B} \right) \times \texttt{depth}_{\texttt{v}_\texttt{B}}$\\
			$\omega_\texttt{B} \leftarrow \left\{\omega_\texttt{B}; \omega_{\texttt{v}_\texttt{B}} \right\}$\\
			$s_{\text d} \leftarrow s_{\text d} - \omega_{\texttt{v}_\texttt{A}} n_{\texttt{v}_\texttt{A}}$\\
		}
		
		
		\BlankLine\BlankLine
        \tcc{Compute virtual impedance (see \sec{reaction_efforts})}
		$f_{\rm r} \leftarrow \frac{\texttt{IMP}_{\texttt{col}}.\texttt{correction}\left( \|{s_\text{d}}\| \right)}{\|{s_\text{d}}\|} s_\text{d}$
		
		\BlankLine\BlankLine
		\tcc{Return forces and torques acting on objects $\mathcal A$ and $\mathcal B$}
		\Return $\left[ \begin{matrix}
					{f}_{\rm r}\\\left({c} - {p}_{\mathcal{A}}\right) \times {f}_{\rm r}
				\end{matrix}\right]$,
				$\left[ \begin{matrix}
					-{f}_{\rm r}\\\left({c} - {p}_{\mathcal{B}}\right) \times -{f}_{\rm r}
				\end{matrix} \right]$
	}
	
	\BlankLine\BlankLine
	\KwResult{
			$\left[ \begin{matrix}
				{f}_{\mathcal{A}}\\{\tau}_{\mathcal{A}}
			\end{matrix}\right]$,
			$\left[ \begin{matrix}
				{f}_{\mathcal{B}}\\{\tau}_{\mathcal{B}}
			\end{matrix} \right]$}
	
	\caption{Computation of reaction efforts magnitude and direction. }
	\label{alg:reaction_efforts}
\end{algorithm}
	\section{Numerical Simulation}\label{sec:simulation}
We use \texttt{Bullet} as a basis for our numerical rigid body simulation. Its modular design has allowed us to quickly replace relevant portions of their solver (pertaining to contact detection and contact efforts) with the proposed approach. In particular, we have replaced the constraint solver used by \texttt{Bullet} with our penalty-based solver (as described in Algorithm \ref{alg:reaction_efforts}). In addition, we replaced the Euler integration with an Ordinary Differential Equation (ODE) solver based on the explicit predictive-corrective Runge-Kutta(4,5) method (\cite{dormand1980family}), \texttt{ode45}, used to integrate \eqn{dyn_trans}--\eqn{dyn_rot}, where efforts $f(t)$, $\tau(t)$, $t\geqslant0$, are composed of reaction efforts provided by \eqn{fr}--\eqn{tr}, gravity's acceleration, and friction forces. To compute the latter, we rely on a Coulomb Friction model adapted from \cite{liu2015experimental}. To circumvent possible issues stemming from the usual discontinuity, we substitute a hyperbolic tangent in the place of the signum function, as described hereafter,
\begin{align}
	f_{\rm f}(t) = -\Big( \mu \|f_{\rm n}(t)\| \tanh \left(\gamma\|\vvR\| \right) + \beta \|v_{\rm r}(t)\|&\Big) v_{\rm rn}(t), \nonumber \\
	t\geqslant0&,
\end{align}
\noindent where $f_{\rm f}(t) \in \mathds{R^3}$ is the friction force (in N), $\mu$, $\beta \in \mathds{R^+}$ represent the Coulomb friction coefficient and the viscous friction coefficient, respectively, $\gamma \in \mathds R^+$ is a scaling factor, $f_{\rm n}(t)\in \mathds{R}^3$ is the reaction force, normal to the surface of contact, $v_{\rm r}(t) \in \mathds{R}^3$ is the relative velocity at contact, and $v_{\rm rn}(t)\triangleq v_{\rm r}(t) /\|v_{\rm r}(t) \|$. The scaling factor $\gamma$ is used to adjust the slope of the zero crossing. In the limit that $\gamma\rightarrow+\infty$, one recovers the usual (discontinuous) model. As in \cite{keller1994}, we define the normal contact force acting on an object as $f_{\rm n}(t) \triangleq f_{\rm r}(t)$, i.e. the previously computed repulsive force.

\subsection{Simulated Grasps}\label{sec:grasp}
\begin{figure}
	\centering
	\scalebox{1}[0.9]{\includegraphics[width=1.0\linewidth]{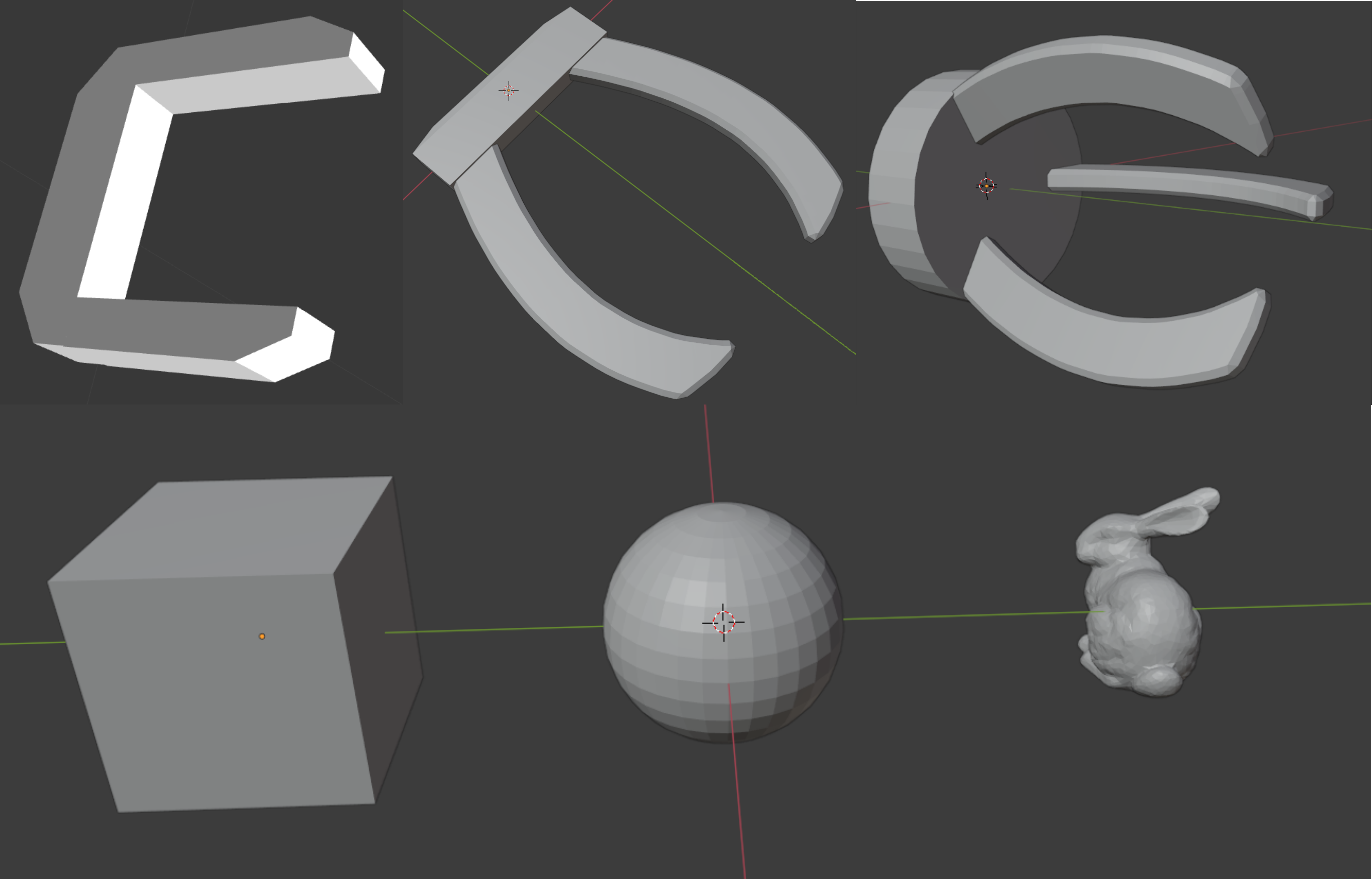}}
	\caption{Grippers (top) and objects (bottom) used for performance evaluation.}
	\label{fig:grippersobjects}
\end{figure}

To evaluate our approach in various grasping scenarii, we have implemented three grippers, representative of a range of structural complexity. The first is a simple 2-fingered prismatic system, the second a 2-fingered revolute gripper, and the final one is a 3-fingered, 3-dimensional revolute gripper. For testing purposes, we have computed the Lagrangian equations of motion for each model, and will be using these instead of the standard constraint-based approach of \texttt{Bullet} to compute the joints' dynamics. Figure \ref{fig:grippersobjects} shows the grippers and objects used for testing.
Three test operations are performed for each gripper individually, with each operation aiming to grasp a particular object. The target objects are a cube, a sphere, and a Stanford Bunny model (\cite{levoy2005scanning}). The mass of each object is set to $0.2$kg. We compare the results of our approach to those obtained with \texttt{Bullet}'s default constraint-based method. Simulation length is chosen to be $10$s, and the following parameters are used, $g_1 = 740 \frac{\text{N}}{\text{cm}}$, $g_2 = 50 \frac{\text{N}}{\text{cms}}$, $g_\text{i} = 250$, $d_\text{t} = 2 \text{mm}$, $d_\text{t} = 2 \text{mm}$, $\mu = \beta = 0.2$, $\gamma = 1$.

\begin{figure}
	\centering
	\scalebox{1}[0.9]{\includegraphics[width=\columnwidth]{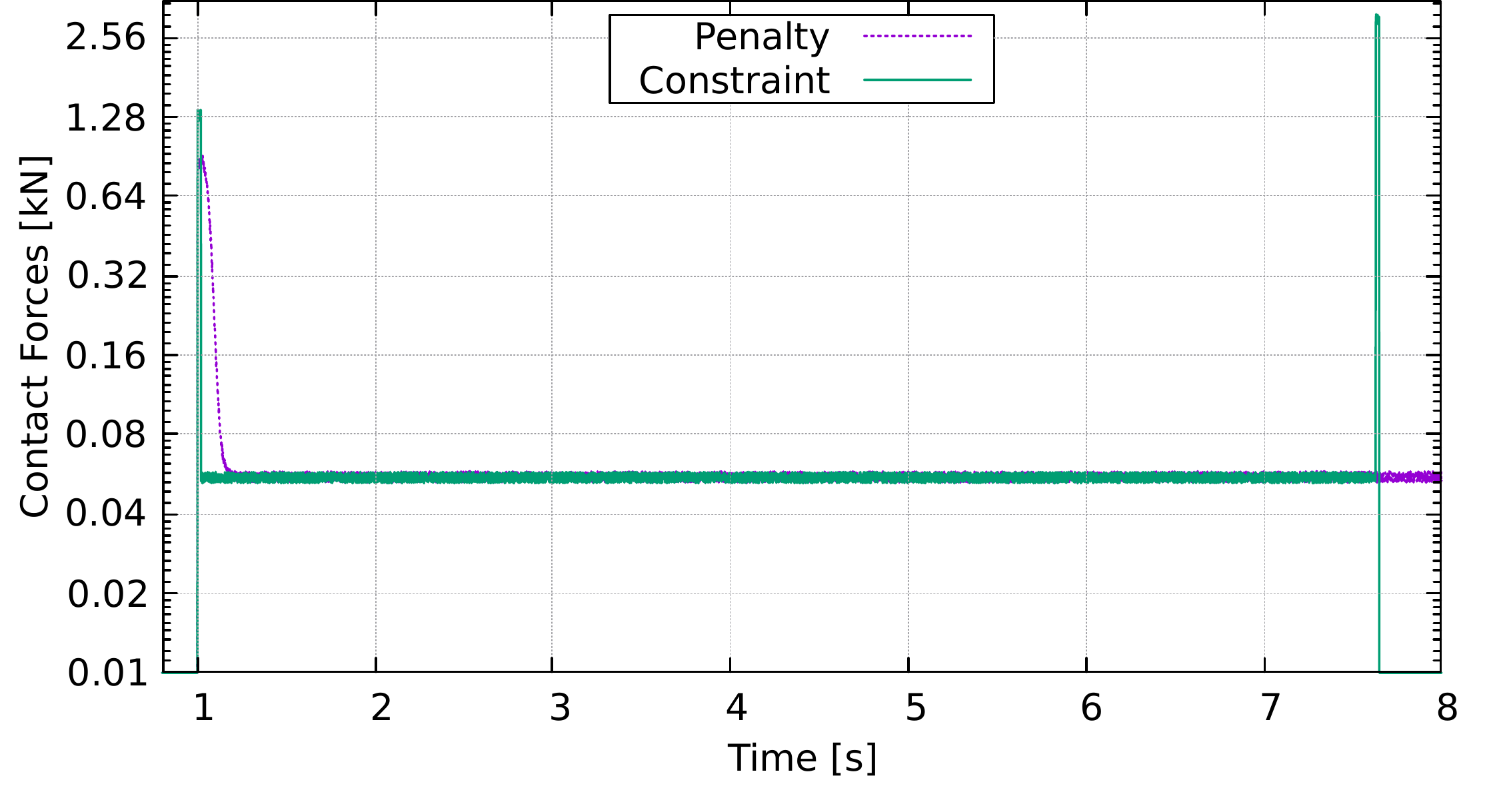}}
	\caption{Grasp failure of constraint-based approach.}
	\label{fig:grasp_explosion}
\end{figure}

The grippers start in an open configuration, the object placed in such a manner that it is ready to be grasped. Once the simulation begins, the fingers enclose the object and exert a constant force from each side for the duration of the simulation. In our experience, using constraint-based approaches has consistently led to failure in finite time for grippers including revolute joints, as shown in \fig{grasp_explosion} where the constraint based approach fails at around simulation time $t=7.7$s (large contact force spike, leading to simulation failure). Here, a 2-finger revolute gripper grasps a ball.

\begin{figure}
	\centering
	\scalebox{1}[1]{\includegraphics[width=\columnwidth]{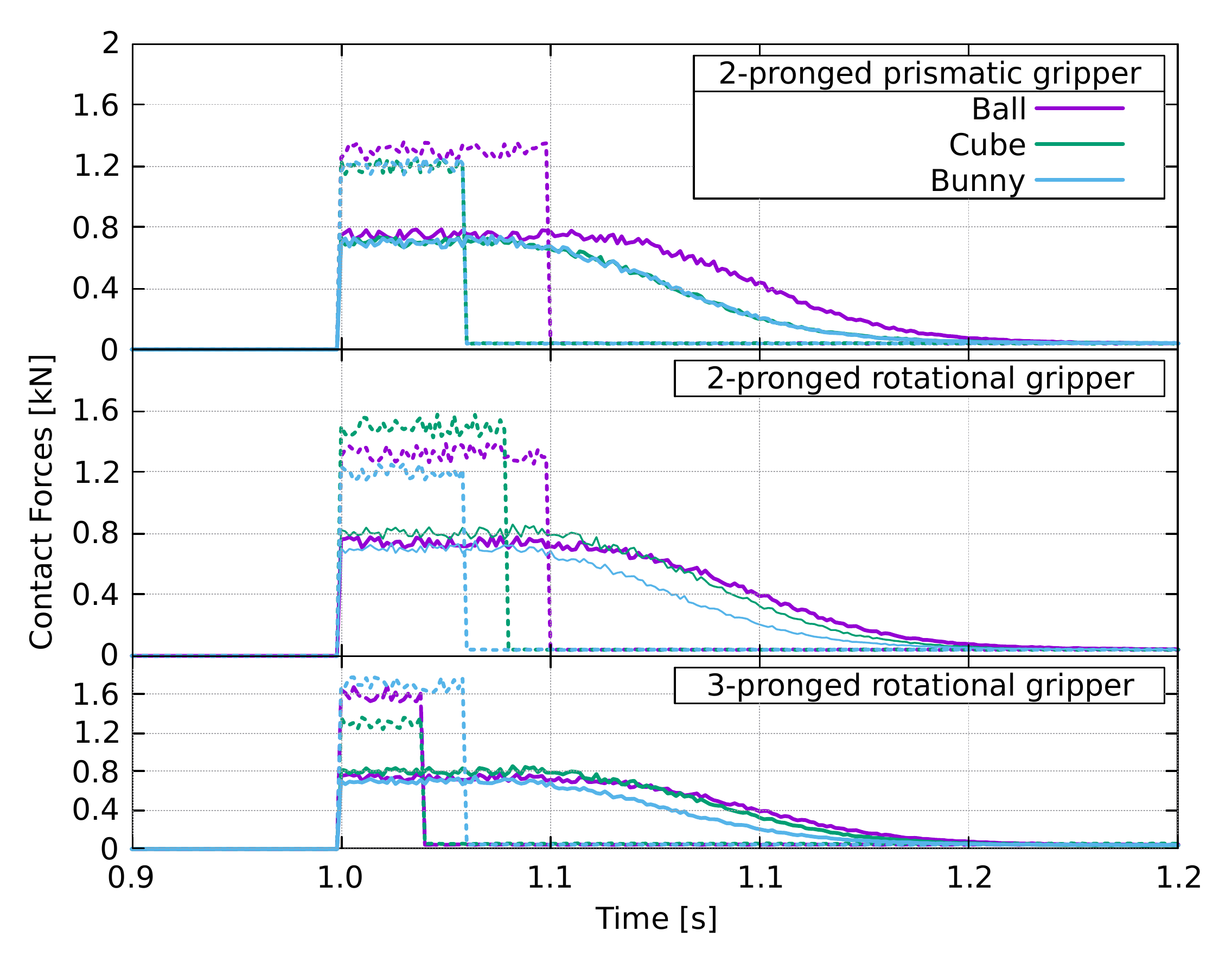}}
	\vspace{-0.7cm}
	\caption{Comparison of contact forces generated by \texttt{Bullet}'s Constraint-based LCP solver and our approach. Dashed lines denote results from the constraint-based simulator, solid lines from the penalty-based method.}
	\label{fig:grasp_throw}
\end{figure}

This behavior does not occur with the proposed approach, which results in the object stably remaining in the gripper's center, between the enclosing fingers. See \fig{grasp_throw} for an illustration of measured contact forces. To better illustrate the shape of the efforts produced upon impact (transient) in comparison to the steady-state value corresponding to contact efforts, we selected a particular time-window.
As shown in \fig{grasp_throw}, the transient efforts upon contact differ between penalty- and constraint-based methods. However, both approaches converge to an identical contact force magnitude once the initial impact has subsided and bodies have reached their steady-state configurations.

	\subsection{Parameter-space exploration}
\label{sec:comparison}
%
To better assess merit of the proposed methodology, we explore grasp behaviors achieved in different areas of the method's parameter space, focusing on parameters which have shown to have the greatest impact on this behavior. For this purpose,
we use the 3-pronged rotational gripper shown in \fig{grippersobjects}, and apply to it a pre-defined sinusoidal pattern. Over the course of this simulation, we measure the kinetic energy of the system, as computed from the objects' simulated linear and angular speeds, and compare it to the expected kinetic energy, as computed from the prescribed movement velocity and assuming no relative movements between object and gripper. In an ideal case, the object would remain at a constant relative position from the gripper, and the ratio of considered kinetic energies would be 1. In situations in which the grasp is less-than-perfect, parasite movements emerge (\emph{wobble creep}), leading to additional (and undesirable) kinetic energy. In the extreme, abrupt large efforts may emerge, disrupting simulation (as is the case when using the constraint method in \fig{grasp_explosion}).

\begin{figure}
	\centering
	\scalebox{0.9}{\includegraphics[width=1.0\columnwidth]{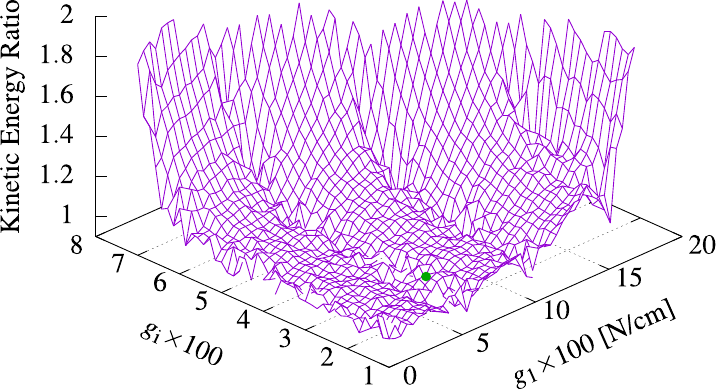}}
	\caption{Kinetic energy ratio comparison for differing values of gain multiplier $g_\text{i}$ and stiffness $g_1$. The green dot indicates the parameters used in \fig{grasp_throw}.}
	\label{fig:impcoeffstiffnesscomp}
\end{figure}

\begin{figure}
	\centering
	\scalebox{0.9}{\includegraphics[width=1.0\columnwidth]{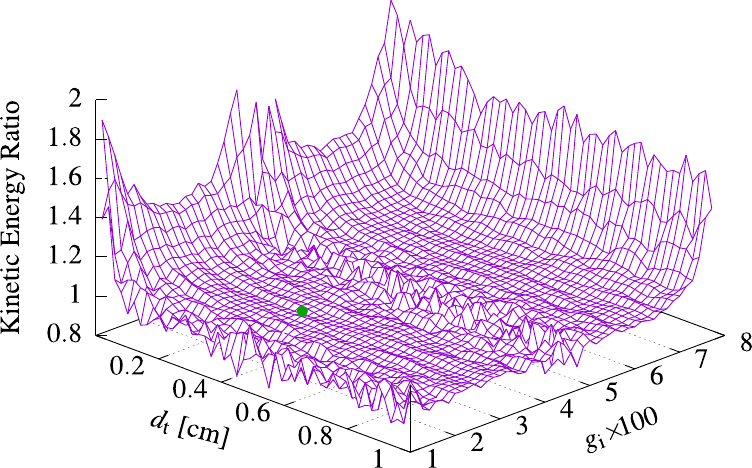}}
	\caption{Kinetic energy ratio comparison for differing values of gain multiplier $g_\text{i}$ and admissible depth $d_\text{t}$. The green dot indicates the parameters used in \fig{grasp_throw}.}
	\label{fig:depthstiffnesscomp}
\end{figure}

As shown in \fig{impcoeffstiffnesscomp}, high contact stiffness $g_\text{1}$ and gain multiplier values $g_\text{i}$ may lead to kinetic energy creation, which implies colored simulation results. Setting these parameters to particularly high values can be verified to result in a jittery behavior for the grasped object, with computed repulsive forces not only reducing contact overlap, but also accelerating objects away from each other.
\fig{depthstiffnesscomp} illustrates the impact on behavior of the depth parameter $d_\text{t}$ and of gain multiplier $g_\text{i}$. Behaviors similar to those observed in the previous example can be seen. Excessively small values of the depth parameter result in a clear increase in kinetic energy over time. It can be observed that decreasing admissible depth leads to greater mechanical stiffness, which in turn leads to a jittery behavior of the grasped object.
%

	\subsection{Numerical Simulation of ShadowRobot Hand}\label{sec:sr_hand}

\begin{figure*}
	\centering
	\scalebox{1}[0.95]{\includegraphics[width=\textwidth,trim={0 3cm 0 7cm},clip]{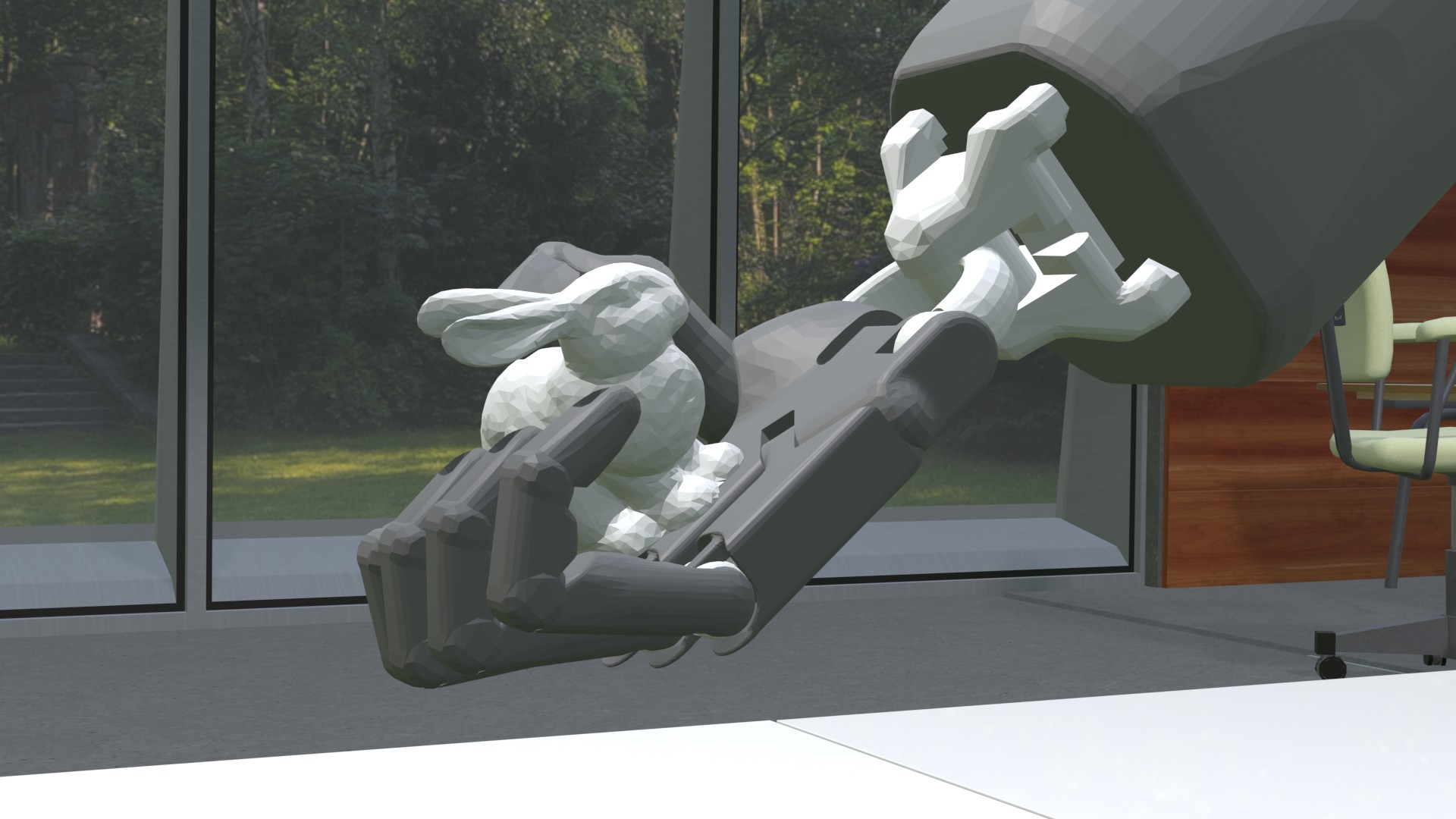}}
	\caption{ShadowRobot Hand grasping the Stanford bunny model (\cite{levoy2005scanning}).}
	\label{fig:srhand}
\end{figure*}

\begin{figure}
	\centering
	\scalebox{1}[0.90]{\includegraphics[width=\linewidth]{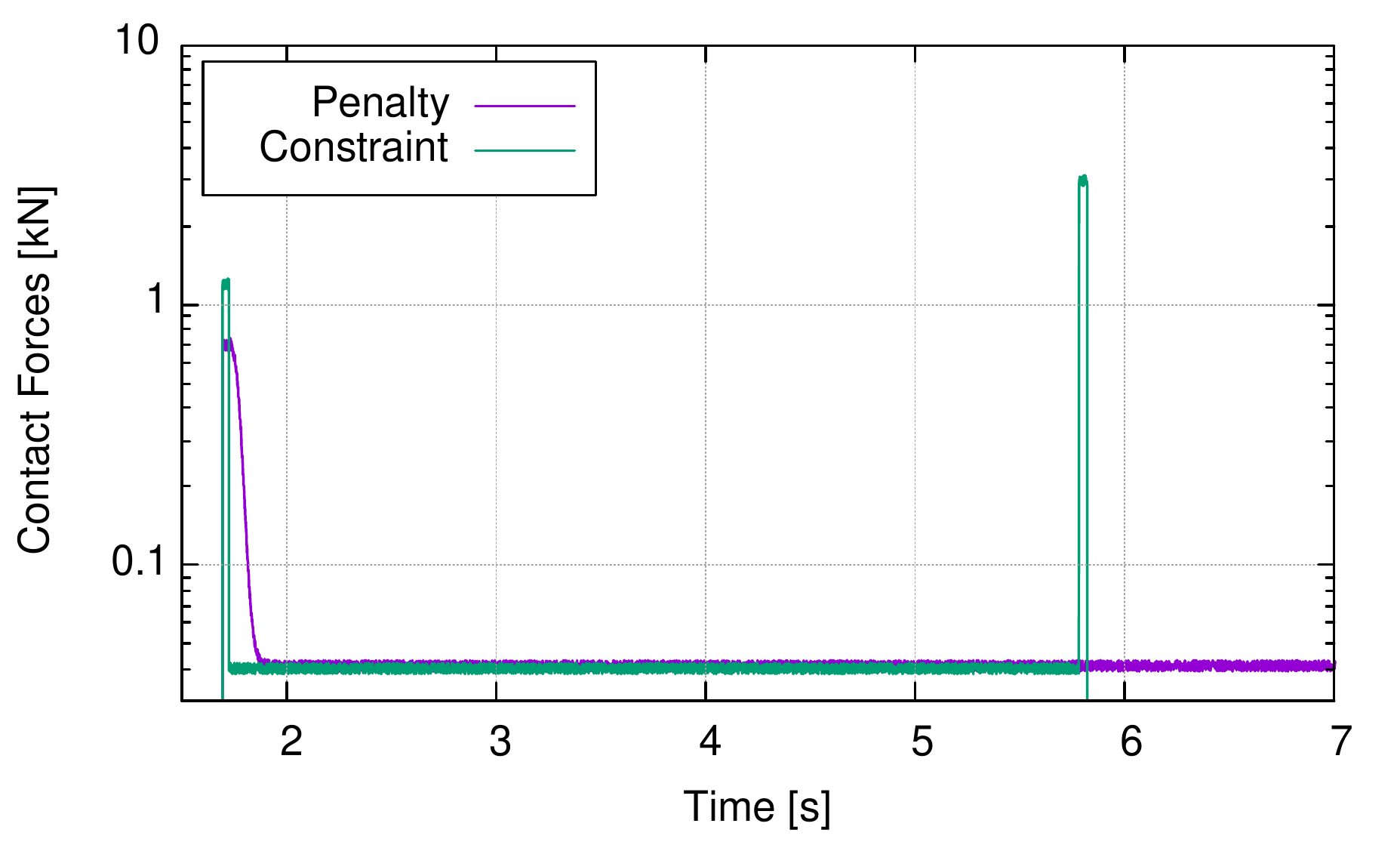}}
	\caption{Contact force comparison for ShadowRobot Hand.}
	\label{fig:srhandcontactforces}
\end{figure}

The grippers used during previous tests have all been developed specifically for the above simulations. They were designed to test the approach's performance under a number of varying circumstances, and compare the results to those obtained using alternate grasping simulation methods. To illustrate efficacy of our approach using a real-world, complex gripper, we have applied our algorithm to a grasping scenario considering the ShadowRobot Hand (see Figure \ref{fig:srhand}). This anthropomorphic gripper features 24 joints and 20 DoFs. As done in previous examples, we compared performance of the proposed approach to that of a constraint-based method. As was the case with simpler grippers, the constraint-based method displays a singular behavior after some simulation time, whereas the proposed penalty-based algorithm behaves as desired. See \fig{srhandcontactforces}, which shows the constraint-based approach failing at about time instant $t=5.8$s.
%
%
	\section{Conclusion}\label{sec:conclusion}
This paper presents a novel penalty-based method to numerically simulate interactions between in-contact rigid bodies, with applications to robotic grasping. A simple, computationally inexpensive method is provided to assess the overlap between objects. The information gathered is in turn exploited to define apposite interaction efforts, designed to ensure the overlap remains a transient occurrence. Contact efforts are computed to offer a reflection of the mechanical stiffness of real-life collisions. Using numerical simulations, we show that the proposed method stably simulates a range of grasping scenarii, including detailed objects and multi-pronged grippers. A comparison to a constraint-based approach, in multiple grasping situations, further illustrates the merit of the proposed approach. In particular it illustrates the fact that, in the simulated scenarii, the proposed method allows to achieve physically plausible results, whereas the constraint-based approach failed to do so.
Work is being invested in integrating the existing approach within the Human Brain Project's NeuroRobotic Platform (NRP, \cite{knoll2016}), to extend its capacities in terms of robust grasping simulation. 
	
	\section{REFERENCES}
	\bibliographystyle{adaptive}
	\begingroup
	\renewcommand{\section}[2]{}%
	\bibliography{IEEEabrv,IEEEexample}

\begin{thebibliography}{10}

\bibitem{schuh2017}
G.~Schuh, R.~Anderl, J.~Gausemeier, M.~Ten~Hompel, and W.~Wahlster, {\em
  Industrie 4.0 Maturity Index: Managing the Digital Transformation of
  Companies}.
\newblock Utz, Herbert, 2017.

\bibitem{haessig1991}
D.~A. Haessig~Jr and B.~Friedland, ``On the modeling and simulation of
  friction,'' 1991.

\bibitem{graspit}
A.~T. Miller and P.~K. Allen, ``Graspit! a versatile simulator for robotic
  grasping,'' {\em IEEE Robotics \& Automation Magazine}, vol.~11, no.~4,
  pp.~110--122, 2004.

\bibitem{openrave}
R.~Diankov and J.~Kuffner, ``Openrave: A planning architecture for autonomous
  robotics,'' {\em Robotics Institute, Pittsburgh, PA, Tech. Rep.
  CMU-RI-TR-08-34}, vol.~79, 2008.

\bibitem{smith2005}
R.~Smith {\em et~al.}, ``Open dynamics engine,'' 2005.

\bibitem{bullet_physics_library}
E.~Coumans {\em et~al.}, ``Bullet physics library,'' {\em Open source:
  bulletphysics. org}, vol.~15, no.~49, p.~5, 2013.

\bibitem{baraff1993non}
D.~Baraff, ``Non-penetrating rigid body simulation,'' {\em SotA reports}, 1993.

\bibitem{cottle2009}
R.~W. Cottle, {\em Linear complementarity problem}.
\newblock Springer, 2009.

\bibitem{anitescu_time_stepping_lcp_solver}
M.~Anitescu, F.~A. Potra, and D.~E. Stewart, ``Time-stepping for
  three-dimensional rigid body dynamics,'' {\em Computer methods in applied
  mechanics and engineering}, vol.~177, no.~3-4, pp.~183--197, 1999.

\bibitem{lemke_lcp_solver}
C.~E. Lemke, ``On complementary pivot theory,'' {\em Mathematics of the
  Decision Science}, vol.~1, pp.~95--114, 1968.

\bibitem{extending_ode_robots}
E.~Drumwright, J.~Hsu, N.~Koenig, and D.~Shell, ``Extending open dynamics
  engine for robotics simulation,'' in {\em International Conference on
  Simulation, Modeling, and Programming for Autonomous Robots}, pp.~38--50,
  Springer, 2010.

\bibitem{engine_simulation}
T.~Erez, Y.~Tassa, and E.~Todorov, ``Simulation tools for model-based robotics:
  Comparison of bullet, havok, mujoco, ode and physx,'' in {\em Robotics and
  Automation (ICRA), 2015 IEEE International Conference on}, pp.~4397--4404,
  IEEE, 2015.

\bibitem{nguyen1988constructing}
V.~Nguyen, ``Constructing force-closure grasps,'' {\em The International
  Journal of Robotics Research}, vol.~7, no.~3, pp.~3--16, 1988.

\bibitem{baraff_analytical}
D.~Baraff, ``Analytical methods for dynamic simulation of non-penetrating rigid
  bodies,'' in {\em ACM SIGGRAPH Computer Graphics}, vol.~23, pp.~223--232,
  ACM, 1989.

\bibitem{taylor2016analysis}
J.~R. Taylor, E.~M. Drumwright, and J.~Hsu, ``Analysis of grasping failures in
  multi-rigid body simulations,'' in {\em IEEE International Conference on
  Simulation, Modeling, and Programming for Autonomous Robots (SIMPAR)},
  pp.~295--301, IEEE, 2016.

\bibitem{yamane2008}
K.~Yamane and Y.~Nakamura, ``A numerically robust lcp solver for simulating
  articulated rigid bodies in contact,'' {\em Proceedings of robotics: science
  and systems IV, Zurich, Switzerland}, vol.~19, p.~20, 2008.

\bibitem{impulse_based}
B.~V. Mirtich, {\em Impulse-based dynamic simulation of rigid body systems}.
\newblock University of California, Berkeley, 1996.

\bibitem{drumwright_penalty_based}
E.~Drumwright, ``A fast and stable penalty method for rigid body simulation,''
  {\em IEEE Transactions on Visualization and Computer Graphics}, vol.~14,
  no.~1, pp.~231--240, 2008.

\bibitem{sagardia_penalty}
M.~Sagardia, T.~Stouraitis, and J.~L. e~Silva, ``A new fast and robust
  collision detection and force computation algorithm applied to the physics
  engine bullet: Method, integration, and evaluation,'' in {\em Conference and
  Exhibition of the European Association of Virtual and Augmented Reality},
  2014.

\bibitem{mirtich1998}
B.~Mirtich, ``Rigid body contact: Collision detection to force computation,''
  in {\em Workshop on Contact Analysis and Simulation, IEEE Intl. Conference on
  Robotics and Automation}, 1998.

\bibitem{stoianovici1996critical}
D.~Stoianovici and Y.~Hurmuzlu, ``A critical study of the applicability of
  rigid-body collision theory,'' 1996.

\bibitem{diolaiti2005}
N.~Diolaiti, C.~Melchiorri, and S.~Stramigioli, ``Contact impedance estimation
  for robotic systems,'' {\em IEEE Transactions on Robotics}, vol.~21, no.~5,
  pp.~925--935, 2005.

\bibitem{tong2004lectures}
D.~Tong, ``Classical dynamics lectures,'' {\em Lagrangian Formalism}, 2004.

\bibitem{jimenez20013d}
P.~Jim{\'e}nez, F.~Thomas, and C.~Torras, ``3d collision detection: a survey,''
  {\em Computers \& Graphics}, vol.~25, no.~2, pp.~269--285, 2001.

\bibitem{jacobson2016libigl}
A.~Jacobson, D.~Panozzo, C.~Sch{\"u}ller, O.~Diamanti, Q.~Zhou, N.~Pietroni,
  {\em et~al.}, ``libigl: A simple c++ geometry processing library,'' 2016.

\bibitem{kagiwada2013numerical}
H.~Kagiwada, R.~Kalaba, N.~Rasakhoo, and K.~Spingarn, {\em Numerical
  derivatives and nonlinear analysis}, vol.~31.
\newblock Springer Science \& Business Media, 2013.

\bibitem{jain2011controlling}
S.~Jain and C.~K. Liu, ``Controlling physics-based characters using soft
  contacts,'' in {\em ACM Transactions on Graphics}, vol.~30, p.~163, 2011.

\bibitem{erickson2003contact}
D.~Erickson, M.~Weber, and I.~Sharf, ``Contact stiffness and damping estimation
  for robotic systems,'' {\em The International Journal of Robotics Research},
  vol.~22, no.~1, pp.~41--57, 2003.

\bibitem{dormand1980family}
J.~R. Dormand and P.~J. Prince, ``A family of embedded runge-kutta formulae,''
  {\em Journal of computational and applied mathematics}, vol.~6, no.~1,
  pp.~19--26, 1980.

\bibitem{liu2015experimental}
Y.~Liu, J.~Li, Z.~Zhang, X.~Hu, and W.~Zhang, ``Experimental comparison of five
  friction models on the same test-bed of the micro stick-slip motion system,''
  {\em Mechanical Sciences}, vol.~6, no.~1, pp.~15--28, 2015.

\bibitem{keller1994}
H.~Keller, H.~Stolz, A.~Ziegler, and T.~Br{\"a}unl, ``Virtual
  mechanics-simulation and animation of rigid body systems,'' 1994.

\bibitem{levoy2005scanning}
M.~Levoy, J.~Gerth, B.~Curless, and K.~Pull, ``{T}he {S}tanford 3{D} {S}canning
  {R}epository.'' \url{ http://www-graphics.stanford.edu/data/3dscanrep}, 2005.
\newblock [Online; accessed 27-02-2021].

\bibitem{knoll2016}
A.~Knoll and M.-O. Gewaltig, ``Neurorobotics: a strategic pillar of the human
  brain project,'' {\em Science Robotics}, pp.~2--3, 2016.

\end{thebibliography}
	\endgroup
	
\end{document}